\documentclass[a4paper, 10 pt, conference]{ieeeconf}  % Comment this line out if you need a4paper

\IEEEoverridecommandlockouts                              % This command is only needed if 
\newcommand{\vx}{\mathbf{x}}
\newcommand{\vy}{\mathbf{y}}

\newcommand{\btheta}{\bm{\theta}}
\newcommand{\loss}{\mathcal{L}}
\newcommand{\mt}{multi-task}

\newcommand\eg{\emph{e.g.}} 
 
\newcommand\etc{\emph{etc}}

\makeatletter
\DeclareRobustCommand\onedot{\futurelet\@let@token\@onedot}
\def\@onedot{\ifx\@let@token.\else.\null\fi\xspace}
\makeatother
\usepackage{times}
\usepackage{epsfig}
\usepackage{graphicx}
\usepackage{amsmath}
\usepackage{amssymb}
\usepackage{textcomp}
\usepackage{multirow}
\usepackage{adjustbox}
\usepackage{dsfont}
\usepackage{booktabs}
\usepackage{bm}
\usepackage{xcolor,soul}

\usepackage{array}
\usepackage{dashrule}
\usepackage{breqn}
\usepackage{colortbl}
\usepackage{hhline}
\usepackage{subcaption}
\usepackage{verbatim}
\usepackage{siunitx}
\usepackage{gensymb}

\title{\LARGE \bf
Dynamic Task Weighting Methods for Multi-task Networks 
\\ in Autonomous Driving Systems
}

\author{
Isabelle Leang$^{1}$, 
Ganesh Sistu$^{2}$, 
Fabian B\"urger$^{1}$, 
Andrei Bursuc$^{3}$ and
Senthil Yogamani$^{2}$ \\ % <-this % stops a space 
$^{1}$Valeo DAR Bobigny, France \hspace{0.3cm}
$^{2}$Valeo Vision Systems, Ireland \hspace{0.3cm}
$^{3}$Valeo.ai, France
} % <-this % stops a space

\begin{document}

\maketitle
\thispagestyle{empty}
\pagestyle{empty}

%%%%%%%%% ABSTRACT
\begin{abstract}
Deep multi-task networks are of particular interest for autonomous driving systems. They can potentially strike an excellent trade-off between predictive performance, hardware constraints and efficient use of information from multiple types of annotations and modalities. However, training such models is non-trivial and requires balancing learning over all tasks as their respective losses display different scales, ranges and dynamics across training. Multiple task weighting methods that adjust the losses in an adaptive way have been proposed recently on different datasets and combinations of tasks, making it difficult to compare them. In this work, we review and systematically evaluate nine task weighting strategies on common grounds on three automotive datasets (KITTI, Cityscapes and WoodScape). We then propose a novel method combining evolutionary meta-learning and task-based selective backpropagation, for computing task weights leading to reliable network training. Our method outperforms state-of-the-art methods by a significant margin on a two-task application.
% \ABc{I would remove the $3\%$. We can just say we outperform by a significant margin.}

% Autonomous driving systems have several visual perception algorithms such as object detection, semantic segmentation, and depth estimation. A joint multi-task learning model has proven to be an effective solution in such scenarios. A main challenge in optimal learning of such a multi-task model is to balance the learning of all the tasks as the loss functions have different scales and ranges. Task weighting methods attempt to solve this problem by weighting the loss functions adaptively. In this paper, we review and systematically evaluate nine task weighting strategies on three automotive datasets namely KITTI, Cityscapes and WoodScape.  
% In particular, we are interested in the behavior of a small efficient network which is deployed on a low-power embedded system. 
% We propose a novel method which is a combination of asynchronous backpropagation and evolutionary Metalearning. Applied on a two task network the proposed method shows a 3\% improvement over the state of the art methods.
\end{abstract}

\section{Introduction}

Visual perception is at the heart of autonomous systems and vehicles \cite{horgan2015vision, heimberger2017computer}. This field has seen tremendous progress during the recent wave of Deep Neural Network (DNN) architectures and methods \cite{ szegedy2014going, he2016deep, he2017mask}. The large majority of computer vision benchmarks are currently dominated by diverse and increasingly effective models encouraging further use in practical applications, \eg,~automatic diagnosis for healthcare, traffic surveillance, autonomous vehicles, \etc.
Such methods reach top performances on individual tasks by leveraging multi-million parameter models requiring powerful hardware, usually for training, but also for predictions.
Perception systems in autonomous vehicles must analyse and understand their surroundings at all time in order to support the multiple micro-decisions needed in traffic, \eg,~steering, accelerating, braking, signaling, \etc. Consequently, a plethora of specific tasks must be addressed simultaneously, \eg,~object detection, 
% semantic 
segmentation~\cite{siam2017deep}, depth estimation~\cite{kumar2018monocular}, motion estimation ~\cite{siam2018modnet,yahiaoui2019fisheyemodnet}, localization~\cite{milz2018visual}, soiling detection~\cite{uvrivcavr2019soilingnet}. Meanwhile hardware constraints in vehicles are limiting significantly DNN capacity and the number of tasks that can be solved. 
% Using a neural network for each individual task is an unfeasible direction.
% \ABc{We should be more clear in why this is unfeasible. I will come back to this depending on space availablity} 
% Thus Multi-Task Learning (MTL) is a highly appealing solution striking a good compromise between the two sides, reliable and high performing methods under limited hardware. 
Using a DNN for each individual task becomes an unfeasible direction.
Thus Multi-Task Learning (MTL) is an appealing solution striking a good compromise over constraints: reliable, high performance, limited hardware.

\begin{figure}[t]
\centering
\includegraphics[width=\columnwidth]{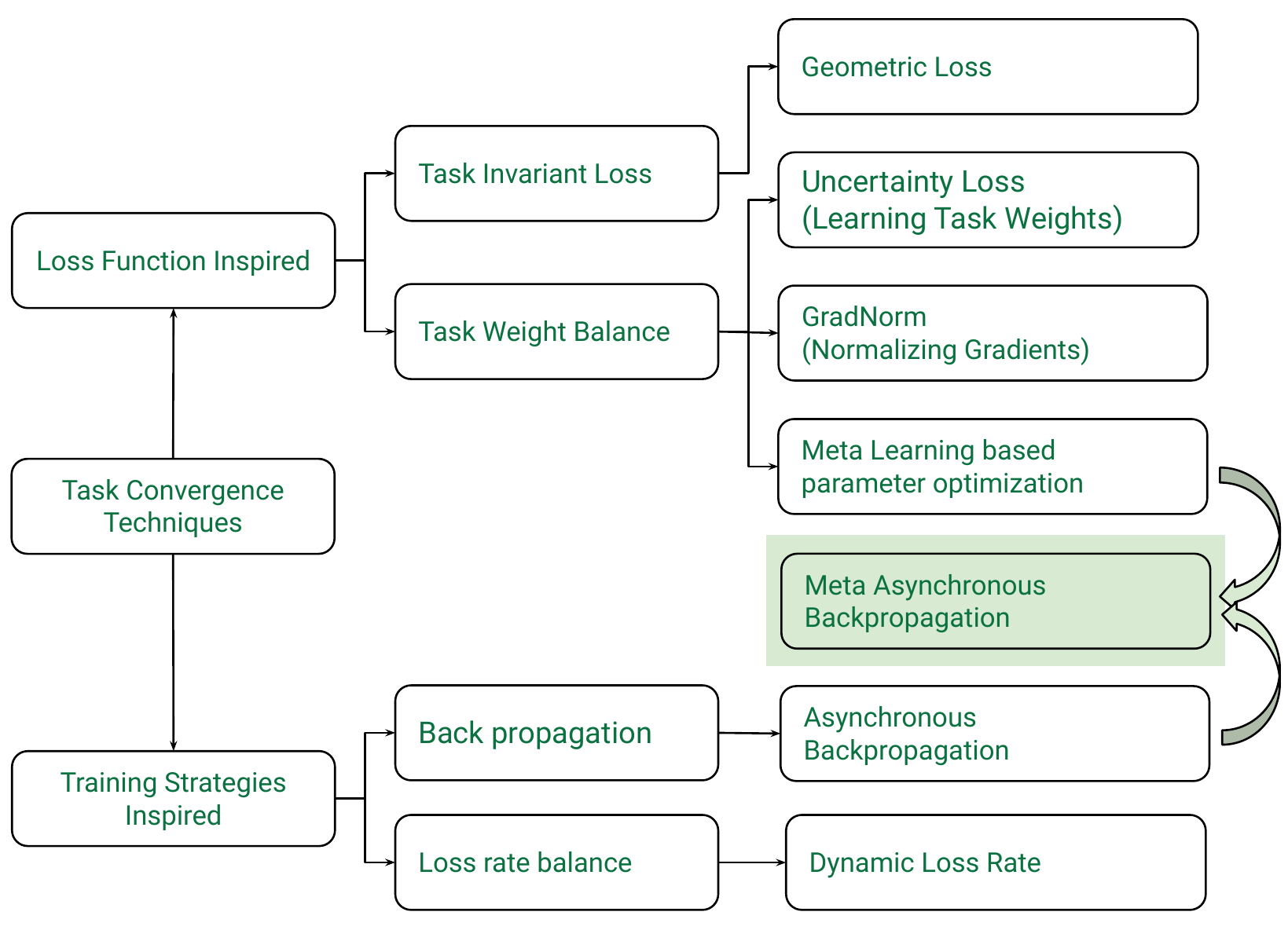}
\caption{Overview of Task Weighting Methods}
\label{fig:TW_methods}
\vspace{-20pt}
\end{figure}

Multi-task networks consist of a shared network backbone followed by a collection of ``heads", typically one for each task. The flexibility of DNNs, makes it easy for practitioners to envision diverse architectures according to the available data and annotations. 
% The main advantage of unified model is improving computational efficiency \cite{sistu2019real,sistu2019neurall}. 
A major advantage of this unified model is computational efficiency~\cite{sistu2019real,sistu2019neurall}. 
% Say there are two problems with two equivalent independent networks utilizing 50\% of available processing power. A unified model with 30\% sharing across the two networks can offer 15\% of additional resources to each network for computing a slightly larger problem. This allows unified models to offer scalability for adding new tasks at a minimal computation complexity. 
Moreover, such models 
% reduce 
save development and training time as shared layers 
% minimize 
replace 
% the need for 
learning of multiple sets of parameters in different models. 
% Unified models learn features jointly for all tasks which makes them robust to over-fitting by acting as a regularizer, as demonstrated in various multi-task networks \cite{kokkinos2017ubernet,neven2017fast, teichmann2018multinet}.
Unified models learn features across tasks, increasing robustness to over-fitting by acting as a regularizer, as shown in previous \mt \ networks~\cite{kokkinos2017ubernet,neven2017fast, teichmann2018multinet}.

However, \mt \ networks are typically difficult to train as different tasks need to be adequately balanced such that learned 
% network 
parameters are useful across all tasks. Furthermore, tasks might have different difficulties and learning paces~\cite{guo2018dynamic} and negatively impact each other once a task starts overfitting before others. Multiple MTL approaches have recently been attempted to mitigate this problem through optimization of multi-task architectures~\cite{ mallya2018packnet, mallya2018piggyback}, learning relationships between tasks~\cite{long2017learning, standley2019tasks} or, most commonly, by weighting the task losses~\cite{Chen2018GradNormGN, kendall2017multi, liu2018endtoend} (Figure~\ref{fig:TW_methods}). 
% Given the versatility of MTL, 
In most works a new problem and task configuration is proposed and only a few baselines are considered. 
% It remains difficult to conclude which technique is better, given a new problem and dataset. 
For a new problem and dataset, it is \emph{a priori} difficult to decide which technique is better.
In this work we benchmark multiple task-weighting methods for a better view on the progress so far.

Meta-learning derived techniques are increasingly popular for solving the tedious and difficult task of tuning hyper-parameters for training a network. Recent methods show encouraging results in finding the network architecture for a given task~\cite{zoph2016neural,liu2019auto}. We propose an evolutionary meta-learning strategy for finding the optimal task weights and exploit our proposed benchmark for emphasizing the interest of such an approach for this problem.

In summary, the contributions of our work are: \textbf{(1)} We conduct a thorough evaluation of several popular and high-performing task-weighting approaches on a two-task setup across three automotive datasets. We notice that among state-of-the-art methods there is no clear winner across datasets as methods are relatively close in performance (including simple baselines) and the ranking is varying. \textbf{(2)} We propose a simple weight learning technique for the two-task setting, where the network learns the task weights by itself. \textbf{(3)} We propose learning the optimal task weights by combining evolutionary meta-learning with task-based selective backpropagation (deciding which tasks to be turned off for a number of iterations). This method outperforms baseline methods across tasks and datasets.   

% \ABc{Ideas left to include:
% \begin{itemize}
%     \item multi task vs multi objective: in MTL tasks are usually complementary
%     % \item MTL are easy to understand but not necessarily easy to deploy
%     % \item task have different degrees of difficulty, and hence different speeds of learning. In addition the difficulty of the task can vary across training
%     \item other difficulties: some tasks might be more compatible than others or even conflicting
%     % \item enumerate contributions and main insights from the current results
%     % \item insight into geometric loss in the log form
%     \end{itemize}}

% \begin{figure*}[htpb]
% \centering
% \includegraphics[width=0.6\textwidth]{fig/TW_methods.pdf}
% \caption{Overview of Task Weighting Methods}
% \label{fig:TW_methods}
% \end{figure*}

% \ABc{We can safely remove this paragraph, given that the paper is short.} Rest of the paper is organized as follows. Section \ref{sec:problem} motivates the problem and describes concrete details of the model which is evaluated. 
% Section \ref{sec:method} discusses the details of the nine weighting strategies used for evaluation. Section \ref{sec:results} presents the results in three different automotive datasets namely KITTI, Cityscapes and WoodScape. Finally, Section \ref{sec:conc} summarizes and concludes the paper.

\section{Related work}

\textbf{Multi-task learning.} MTL is not a novel problem and has been studied before the deep learning revival~\cite{caruana93multitasklearning}. MTL has been applied to various applications outside computer vision, \emph{e.g.}~natural language processing~\cite{collobert2008unified}, speech processing~\cite{huang2015rapid}, reinforcement learning~\cite{lazaric2010bayesian}. For additional background on MTL we refer the reader to this recent review~\cite{ruder2017learning}.

\textbf{Multi-task networks.} In general, MTL is compatible with several computer vision problems where the tasks are rather complementary and help out optimization. MultiNet~\cite{teichmann2018multinet} introduces an architecture for semantic segmentation, object detection and classification. With UberNet~\cite{kokkinos2017ubernet}, Kokkinos tackles 7 computer vision problems over the same backbone architecture. CrossStich networks~\cite{Misra_2016} learn to combine \mt \ neural activations at multiple intermediate layers. Progressive Networks~\cite{rusu2016progressive} consist of multiple sequentially added neural networks with new tasks, and transfer knowledge from previously trained networks to the newly added one (previous networks are frozen each time a new task and network are added). 
% In PackNet~\cite{mallya2018packnet}, the authors train a network over a sequence of tasks and for each new task they train only the least-active neurons from the previous task.
In PackNet~\cite{mallya2018packnet}, a network is trained over a sequence of tasks and for each new task only the least-active neurons are trained selectively.
AuxNet \cite{Chennupativisapp19} uses auxiliary tasks to improve the performance of the main task using multi-task learning.
Rebuffi~\emph{et al}.~\cite{rebuffi2017learning} train a network over 10 datasets and tasks, and for each task require a reduced set of parameters attached to several intermediate layers. In some cases, a single computer vision problem can be transformed into a MTL problem, \eg, Mask R-CNN~\cite{he2017mask} decomposes instance segmentation into object detection + classification + semantic segmentation. This approach is common in object detection~\cite{redmon2016you}.

\textbf{Task loss weighting.} Initial Deep MTL networks made use of a weighted  sum of individual task losses~\cite{teichmann2018multinet,8100062}. Recently, more complex heuristics have started to emerge for  balancing the task weights using: per-task uncertainty estimation~\cite{kendall2017multi}, difficulty of the tasks in terms of precision and accuracy ~\cite{guo2018dynamic}, statistics from task losses over time~\cite{liu2018endtoend} or from their corresponding gradients~\cite{Chen2018GradNormGN}.

\textbf{Meta-learning} is a learning mechanism that uses experience from other tasks. The most common use-case of meta-learning is the automatic adaptation of an algorithm for a task at hand. More specifically, meta-learning can be used for hyper-parameter optimization~\cite{li2016hyperband}, for exploring network architectures~\cite{zoph2016neural,liu2019auto,jaderberg2017population} or various non-trivial combinations of variables, \emph{e.g.}~data augmentation~\cite{cubuk2018autoaugment}. In this line of research, we adapt an evolutionary meta-learning strategy for finding the optimal task weights along with the strategy {for asynchronously training the two tasks.}

\section{Background} \label{sec:problem}

% \begin{figure*}[htpb]
% \centering
% \includegraphics[width=0.6\textwidth]{fig/mtl_arch.pdf}
% \caption{Multi-task visual perception network architecture}
% \label{fig:multi-task}
% \end{figure*}

In the following, we provide a formal definition of the MTL setting which will allow us to provide a common background and easier understanding of the multiple task weighting approaches compared and proposed in this work. Consider an input data space $\mathcal{X}$ and a collection of $T$ tasks $\mathcal{T}=\{\tau_{1:T}\}$ with corresponding labels $\{ \mathcal{Y}_{1:T}\}$.
% $\{ \mathcal{Y}^t\}_{t=1:T}$.
% , where $T$ is the number of tasks
In MTL problems, we have access to a dataset of $N$ i.i.d. samples $\mathcal{D}=\{ \vx^i, \vy^i_1, \dots, \vy^i_T \}$, where $\vy^i_{\tau}$ is the label of the data point $\vx^i$ for the task $\tau$. In computer vision $\vx^i$ usually corresponds to an image, while $\vy^i_{\tau}$ can correspond to a variety of data types, \eg, scalar(s), class label, 2D heatmap/class map, \etc.
% , list of 2D/3D coordinates

The main component in any MTL is a model 
$f\left( \vx; \btheta \right) : \mathcal{X} \rightarrow \{ \mathcal{Y}_{1:T}\}$,
% $f\left( \vx; \btheta \right) : \mathcal{X} \rightarrow \{ \mathcal{Y}^t\}_{t=1:T}$,
which in our case is a CNN with learnable parameters $\btheta$. The most commonly encountered approach for MTL in neural networks is hard parameter sharing~\cite{caruana93multitasklearning}, where there is a set of hidden layers shared between all tasks, \emph{i.e.}, \emph{backbone}, to which multiple task-specific layers are connected. Formally, the model $f$ becomes:
% \begin{equation}
% f\Big( \vx;\btheta^{shared}, \{ \btheta^t \}_{t=1:T} \Big) : \mathcal{X} \rightarrow \{ \mathcal{Y}^t\}_{t=1:T}
% \end{equation}
\begin{equation}
f\Big( \vx;\btheta_{shared}, \{ \btheta_{1:T} \} \Big) : \mathcal{X} \rightarrow \{ \mathcal{Y}_{1:T}\}
\end{equation}

For clarity, we denote as $\{ \bm{\theta}_{\mathcal{T}} \}$ the set of parameters coming from all task-specific layers $\bm{\theta}_{\tau}$. Each task has its own specific loss function $ \loss_{\tau} \Big( f \left( \vx^i; \bm{\theta}_{shared}, \bm{\theta}_{\tau} \right), \vy^i_{\tau}\Big)$ attached to both its specific layers $\bm{\theta}_{\tau}$ and the common backbone $\btheta_{shared}$.
The optimization objective for $f$ boils down to the joint minimization of all the $T$ task losses as following:
% \begin{equation}
%     \min_{\bm{\theta}^{shared}, \{ \bm{\theta}^t \}_{t=[1, T]}} \mathcal{L}_{total}(\bm{\theta}^{shared}, \{ \bm{\theta}^t \}_{t=[1,T]}) = 
%     \sum_{t=1}^T w^t\mathcal{L}^t\Big(f(\mathcal{D}; \bm{\theta}^{shared}, \bm{\theta}^t)\Big) 
% \end{equation}
\begin{multline}
    \min_{\btheta_{shared}, \{ \theta_{\mathcal{T}} \}} \mathcal{L}_{total}(\bm{\theta}_{shared}, \{ \theta_{\mathcal{T}} \}) = \\ \min_{\btheta_{shared}, \{ \btheta_{\mathcal{T}} \}} \sum_{\tau=1}^T w_{\tau}\loss_{\tau}\Big(f(\mathcal{D}; \btheta_{shared}, \btheta_{\tau})\Big) 
\end{multline}
where $w_{\tau}$ are per-task weights that can be static, computed dynamically or learned by $f$, in which case $w_{\tau} \subset \btheta_{\tau}$.

Weighted losses for MTL are intuitive and easy to formulate, however they are more difficult to deploy. The main challenge is related to computing $\{w^{\tau}\}$. This is non-trivial as the optimal weights for a given task can evolve in time depending on the difficulty of the task and of the content of the train set~\cite{Chen2018GradNormGN, guo2018dynamic}, \eg, diversity of samples, class imbalance, \etc. Also, the task weights can depend on the affinity between the considered tasks~\cite{zamir2018taskonomy} and the way 
% the help, 
they complement~\cite{standley2019tasks} or counter each other~\cite{sener2018multitask}, relationships that potentially evolve across training iterations. Recent moment-based optimization algorithms with adaptive updates, SGD with momentum, and adaptive step-size, \eg, ADAM~\cite{kingma2014adam}, can also influence the dynamics of the MTL, by attenuating the impact of a wrongly tuned weight or on the contrary by keeping the bias of a previously wrong direction active for more iterations.
In practice, this challenging problem is solved via lengthy and expensive grid search or alternatively via a diversity of heuristics with varying degrees of complexity. {In this work, we rather explore the former type of approaches and propose two heuristics for estimating optimal weights to improve performances namely simple dynamic task weighting loss approaches and a meta-learning based approach with asynchronous backpropagation.}
\vspace{-20pt}

%\subsection{Motivation for adaptive Loss combination}

% \subsection{An efficient two-task baseline}

% \textcolor{red}{
% \textbf{Network Architecture}
% In this section, we report results on a baseline network design which we plan to improve upon. We propose a jointly learnable shared encoder network architecture provided in the high level block diagram in Figure \ref{fig:multi-task}. We implemented a two task network with 3 segmentation classes (background, road, sidewalk) and 3 object classes (car, person, cyclist). To enable feasibility on a low power embedded system, we used a small encoder namely Resnet10 which is fully shared for the two tasks. FCN8 is used as the decoder for semantic segmentation and YOLO is used as the decoder for object detection. Loss function for semantic segmentation is the cross entropy loss to minimize misclassification. For geometric functions, average precision of object localization is used as error function in the form of squared error loss. We use a weighted sum of individual losses $L = w_{seg}*L_{seg} + w_{det}*L_{det}$ for the two tasks. In case of fisheye cameras which have a large spatially variant distortion, we implemented a lens distortion correction using a polynomial model. 
% }
\section{Task-weighting Methods} \label{sec:method}

In this section, we first review the most frequent task weighting methods encountered in literature and in practice ($\mathcal{x}$\ref{sec:baselines}), and then describe our contributed approaches for this problem ($\mathcal{x}$\ref{sec:weight_learning}, $\mathcal{x}$\ref{sec:metalearning}, $\mathcal{x}$\ref{sec:async_backprop}). 
% In this work 
Here we consider a two-task setup, where we train a CNN for joint object detection and semantic segmentation (Figure~\ref{fig:multi-task}). 
% \ABc{We could specify here why we go for the two-task setup. The ICRA reviewers did not seem to understand why.} 
In the following we will adapt the definitions of the task weighting methods to this setup with $\mathcal{T}=\{det, seg\}$. 
\subsection{Baselines}\label{sec:baselines}
\subsubsection{No task weighting}
An often encountered approach in MTL is to not assign any weights to the task losses~\cite{teichmann2018multinet, neven2017fast, 8100062}. The optimized loss is then just the 
% standard 
sum of individual task losses with all task weights set to 1.0. This can 
% happen 
occur also when the practitioner adds an extra-loss at the output of the network, not necessarily realising that the problem has become MTL. 
While very simple, there are a number of issues with this approach. First the network is now extremely sensitive to imbalances in task data, task loss ranges and scales (cross entropy, $L_2$, \emph{etc}). Due to these variations and desynchronization, some of the task losses advance faster than the others. 
% These task will be reaching overfitting, by the time the other task losses converge, highlighting the necessity of balancing the losses during training. 
Consequently by the time the ``slower" task converges, the ``faster" task will have already overfitted. This highlights the necessity of balancing losses during training.

\subsubsection{Handcrafted task weighting}
% We try to find the best weighting manually. This is possible by testing several values for example. But here, we weigh the task losses so that they fit the same scale: the scale is computed using the task loss values at first iteration and is constant along the training.
Here, the loss weights are found and set manually. We can achieve this by inspecting the value of the loss for several samples. Then the losses are weighted such that they are brought to the same scale: this is computed using the values of the loss at first iterations and remains constant during the training.\footnote{A more technically sound way of selecting the losses  would be to look at the gradients of the losses instead of the values of the losses. However, we include this baseline as it is frequently performed by practitioners when tuning hyper-parameters after short trials and inspections.}
\begin{align}
&\loss_{total} = w_{seg} \loss_{seg} + w_{det} \loss_{det}\\
&w_{seg} = \loss_{det}^{(t=0)}/ \loss_{seg}^{(t=0)}, &w_{det} = 1.0 
\end{align}
where $w_{seg}$ and $w_{det}$ are the weights, $\loss_{seg}$ and $\loss_{det}$ the losses for the semantic segmentation branch and object detection respectively, while $\loss_{\tau}^{(t=0)}$ is the loss for task $\tau$ at the first training iterations.
% $t=0$.
\subsubsection{Dynamic task loss scaling}
% In this method, we try to take into account the dynamic evolution of the task losses during training. We scale the task losses dynamically.
For this method, we take into account the evolution of per-task losses during training. We compute task weights dynamically, at the end of every training epoch as follows:
\begin{align}
&\loss_{total}^{(t)} = w_{seg}^{(t)} \loss_{seg}^{(t)} + w_{det}^{(t)} \loss_{det}^{(t)}\\
&w_{seg}^{(t)} = \tilde{\loss}_{det}^{(t-1)}/\tilde{\loss}_{seg}^{(t-1)},
&w_{det}^{(t)} = 1.0
\end{align}
where $\tilde{\loss}_{\tau}^{(t-1)}$ is the average $\tau$ loss over the previous epoch.
% \ABc{@Isabelle: is the formula correct $w_{seg}^{(t)} = \loss_{det}^{(t-1)}\loss_{seg}^{(t-1)}$?  Is this product of losses derived from the geometric loss? Was is proposed in a previous paper? What is the intuition behind making a product?}

\subsubsection{Uncertainty-based weighting} 
% Kendall~\etal~\cite{kendall2017multi} proposed a loss function based on maximising the Gaussian likelihood with homoscedastic uncertainty. The central idea behind this method and weight learning method is that there are no optimal set of weights for a multitask learning model and hence these weights can be adjusted during the training phase. The main difference between Kendall's and Weight Learning method is that Kendall's work imposes Gaussian likelihood over the parameters.
Kendall~\emph{et al.}~\cite{kendall2017multi} propose looking into aleatoric
% or data 
uncertainty for computing the task weights adaptively during training. They argue that each task has its own homoscedastic uncertainty $\sigma_{\tau}$ which can be learned by the network for each task during training ($\sigma_{\tau} \subset \btheta_{\tau}$). Since they are based on homoscedastic uncertainty, the task weights are not input-dependent and 
% have been shown to 
converge to a constant value after 
% some 
a number of iterations~\cite{kendall2017multi}. 
% The loss functions for this method are derived from the Gaussian likelihood.
The Gaussian likelihood is used as loss function for this method.

\subsubsection{GradNorm} 
This method from~\cite{Chen2018GradNormGN} 
% sees
views \mt \ network training as a problem of unbalanced gradient magnitudes back propagated through the shared layers (encoder). 
% And proposes a solution to normalize the unbalanced task gradients by optimizing a new gradient loss that controls the task loss weights. 
This solution normalizes the unbalanced task gradients by optimizing a new gradient loss that controls the task loss weights. 
% These task 
Task loss weights are updated using gradient descent of this new loss. 
% In all our experiments, the relative inverse training rate of each task is set to 1.

\begin{figure}[tpb]
\centering
\includegraphics[width=\columnwidth]{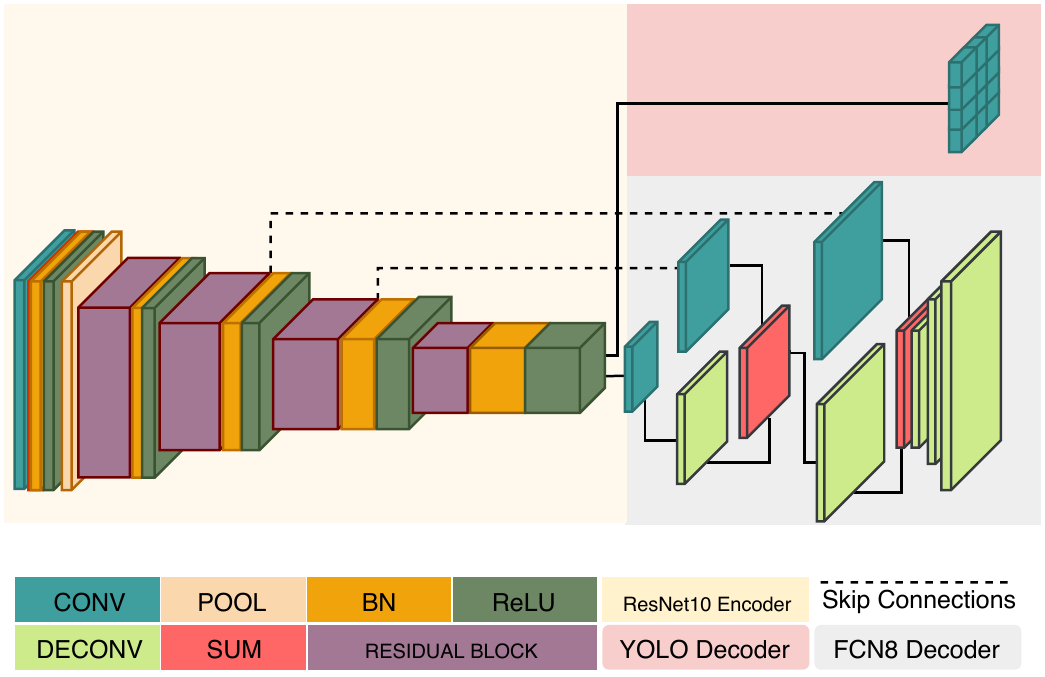}
\caption{Multi-task visual perception network architecture}
\label{fig:multi-task}
\vspace{-20pt}
\end{figure}

% \subsection{Geometric loss} 
\subsubsection{Geometric loss} 
% In \cite{chennupati2019multinet++} authors proposed a parameter free loss function called Geometric Loss Strategy to over come the manual fine tuning of task weights. A geometric mean of losses is used instead of weighted arithmetic mean. For example a $T$ task loss function can be expressed as,
The Geometric Loss Strategy~\cite{chennupati2019multinet++} is a parameter free loss function for overcoming the manual fine tuning of task weights. It consists of a geometric mean of losses instead of the usual weighted arithmetic mean. For example a $T$ task loss function can be expressed as,
% \begin{equation}
% \label{eq:ALS-3task}
%     \loss_{total} =  \sqrt[3]{\loss_{1}\loss_{2}\loss_{3}}
% \end{equation}
\begin{equation}
\label{eq:ALS-3task}
    \loss_{total} =  (\prod_{\tau=1}^{T} \loss_{\tau})^{\frac{1}{T}} 
\end{equation}
The loss strategy was tested with a three task network on KITTI~\cite{Geiger2012CVPR} and Cityscapes~\cite{Cordts2016Cityscapes} datasets. The loss function acts as a dynamically adapted weighted arithmetic sum in log space, these weights act as regularizers and control the rate of convergence between the losses. 
% \ABc{Could be nice to explain what is expected to happen here with the three losses. Eventually looking at this loss after applying log and seeing as a sum, might be interesting for insight}
\\

In the following we describe our proposed approaches for task weighting.

\subsection{Weight learning}\label{sec:weight_learning}
% In \cite{doersch2017multi} cross connections between a shared encoder and task specific decoder are adjusted as learnable parameters. In~\cite{kendall2017multi} task weighting parameters are learned during the training. Inspired by these two works we propose a single parameter learning strategy for a two task network as follows, 
Doersch and Zisserman~\cite{doersch2017multi} use weighted cross connections between the shared encoder and task specific decoders, adjusted via learning. In~\cite{kendall2017multi} task weighting parameters are learned during training. Inspired by these two approaches, we propose a single parameter learning strategy for a two-task network as follows:
% \begin{align}
% &\loss_{total}^{(t)} = \alpha^{(t)} \loss_{seg}^{(t)} + (1-\alpha^{(t)}) \loss_{det}^{(t)} \\
% &\alpha^{(t)} = \text{Sigmoid}(w^{(t)}) %\text{ } \forall t \\ 
% \end{align}
\begin{align}
&\loss_{total} = \alpha \loss_{seg} + (1-\alpha) \loss_{det} \\
&\alpha = \text{Sigmoid}(w_{(\btheta_{shared}, \btheta_{\mathcal{T}})}) %\text{ } \forall t \\ 
\end{align}
where $\alpha$ is the weight balancing term 
% and it is 
computed from the learnable parameter $w_{(\btheta_{shared}, \btheta_{\mathcal{T}})} \subset \{\btheta_{shared}, \btheta_{\mathcal{T}} \}$, which is updated by backpropagation at each training iteration.
Note that here the task weights are updated after each mini-batch. 

This simple weight learning method enables the network to adjust by itself the pace of learning of the two tasks. 
The sigmoid outputting the $\alpha$ term serves as a gating mechanism~\cite{cho2014learning} to balance the two tasks while taking into 
% consideration 
account the interactions between the two. Bounding the weights in $[0,1]$ implicitly regularizes learning by removing the risk of having extremely unbalanced task weights. 
% \ABc{The following line can be commented out in case of space problems.}
% The advantage of this method on the two-task setting compared to the uncertainty-based one~\cite{kendall2017multi} is that this is a line search (only one parameter to adjust) whereas the latter is a 2D space search for a two task network.

% \ABc{It is not clear what is meant in the following sentence: The advantage of this method compared to Kendall's uncertainty is that this is a line search where as later method is a 2D space search for a two task network.}

\subsection{Task weighting using Evolutionary Meta-learning}\label{sec:metalearning}

% The task weighting problem can be understood 
% viewed as a hyperparameter optimization problem with 
% % $n_{tasks}$ 
% $T$ numeric variables equal to the number of tasks. An efficient and extended version of Evolution Strategies~\cite{rechenberg1978} is used as base optimization method. 
Task weighting can be understood 
viewed as a hyper-parameter optimization problem with 
$T$ numeric variables equal to the number of tasks. We use as base method an efficient and extended version of Evolution Strategies~\cite{rechenberg1978} (ES). The extensions of ES allow the optimization of linearly and exponentially scaled numerical variables as well as categorical variables simultaneously \cite{burger2016understanding}. All variables are treated in an independent way so that the system can handle any number of variables. Furthermore, the variable gradient information is exploited in a semi-greedy way in the mutation operation which is inspired from Natural Evolution Strategies~\cite{wierstra2014}. The gradient towards the last most promising direction with respect to the target metric is added as a bias for every numerical value. Together with the random noise of the mutation, the algorithm can escape local minima while converging fast.  Finally, in order to prevent repeated evaluations of the same region in the search space, a Tabu search method \cite{glover1986} is applied. A history of all tested configurations is stored and a distance metric between them is defined for all numerical variables with respect to the relative differences normalized to the search space range. The mutation operation then generates candidates that have to fulfill a minimum distance of at least 0.1\% of the search space range towards already tested solutions.

The search space is defined as numerical variable for each task
% $w_{\tau_i}' = 10^{w_{\tau_i}}$ with $w_{\tau_i}' \in [min_{\tau_i},max_{\tau_i}]$. 
$\tau \in \mathcal{T}$ as $w_{\tau}' = 10^{w_{\tau}}$ with $w_{\tau}' \in [min_{\tau},max_{\tau}]$.
The weight is optimized on an exponential scale as the optimal weight ratio can be non-linear. Furthermore, the final task weight coefficients are normalized such that their sum is one with the goal to leave the overall magnitude in the loss unchanged, \emph{i.e.}~$\bar{w}_{\tau} =  \frac{w_{\tau}'}{\sum^{T}_{j=1}w_{j}'}$.
% $w_{task,i,norm} =  \frac{w_{task,i}'}{\sum^{n_{tasks}}_{j=1}w_{task,j}'}$.

% \ABc{@Fabian: It's not clear here what is the index $i$ referring to. Is it the task index? We might need to provide a bit more details about what is happening here. The reviewers will be most likely with more robotics background than ML and they might not understand what you are doing here.}

% In order to guide the optimization to an equilibrium between the tasks, the geometric mean between the detection 
% % mean average precision
% mAP and the segmentation 
% % intersection over union 
% mIoU is used as target metric. 
% % \ABc{This part could go in the implementation details in section V to avoid cluttering here: }

In order to guide the optimization to an equilibrium between the tasks, the geometric mean between the detection  mAP and the segmentation mIoU is used as target metric. 

% A dynamic weight transfer is used that reuses the currently best models weights during the training. So the number of epochs for each run can be reduced to 8 epochs (for Woodscape dataset) to do continuously finetuning while simultaneously tuning the hyperparameters.

\begin{figure}[tpb]
\centering
\begin{subfigure}{0.45\textwidth}
\centering
\includegraphics[width=0.7\linewidth,keepaspectratio]{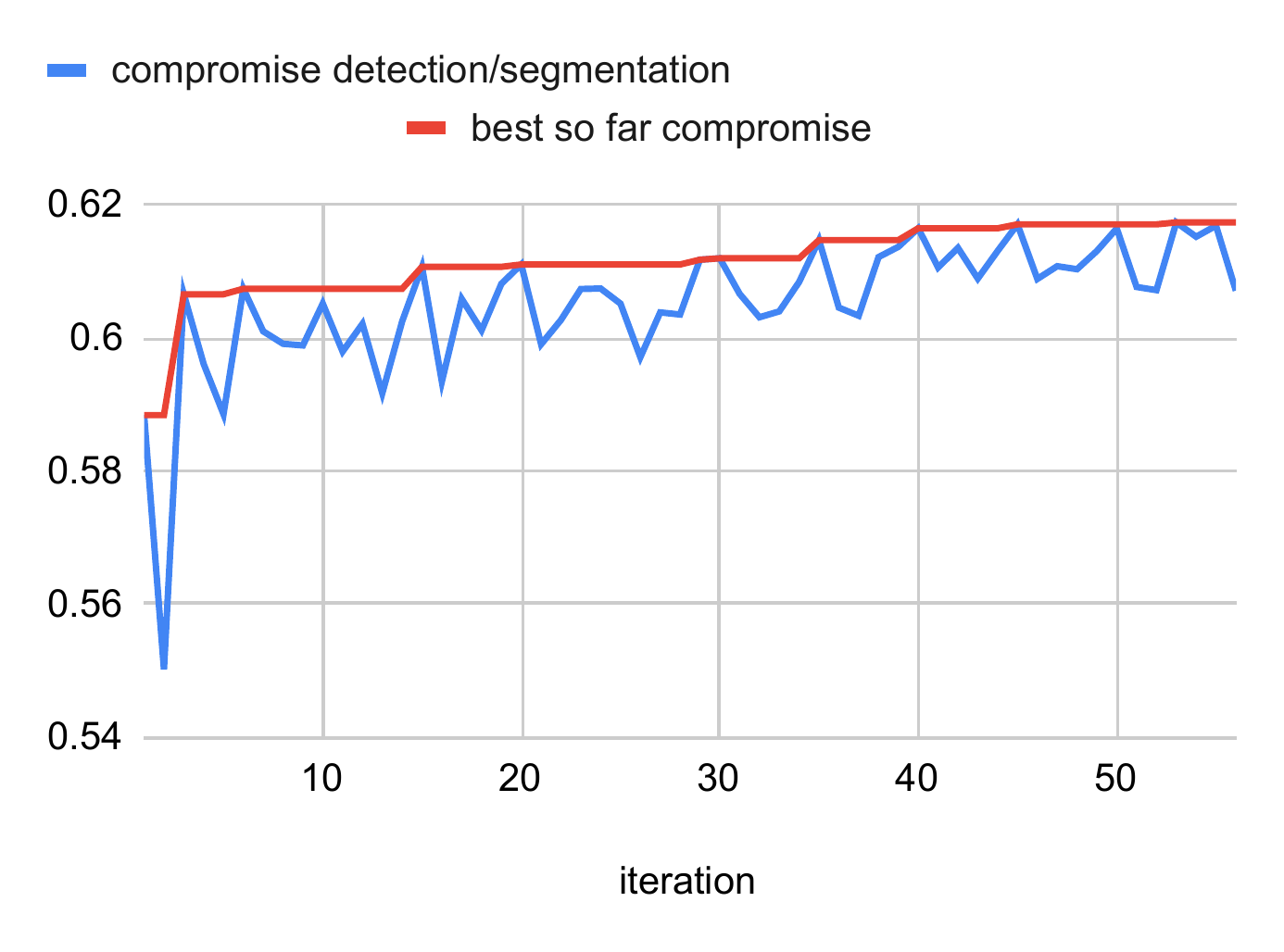}
\caption{Geometric mean of mIoU and mAP over time.}
\label{fig:woodscape_metalearning_async_backprop_main_metric}
\end{subfigure}
\hfill
\begin{subfigure}{0.4\textwidth}
\centering
\includegraphics[width=0.7\linewidth,keepaspectratio]{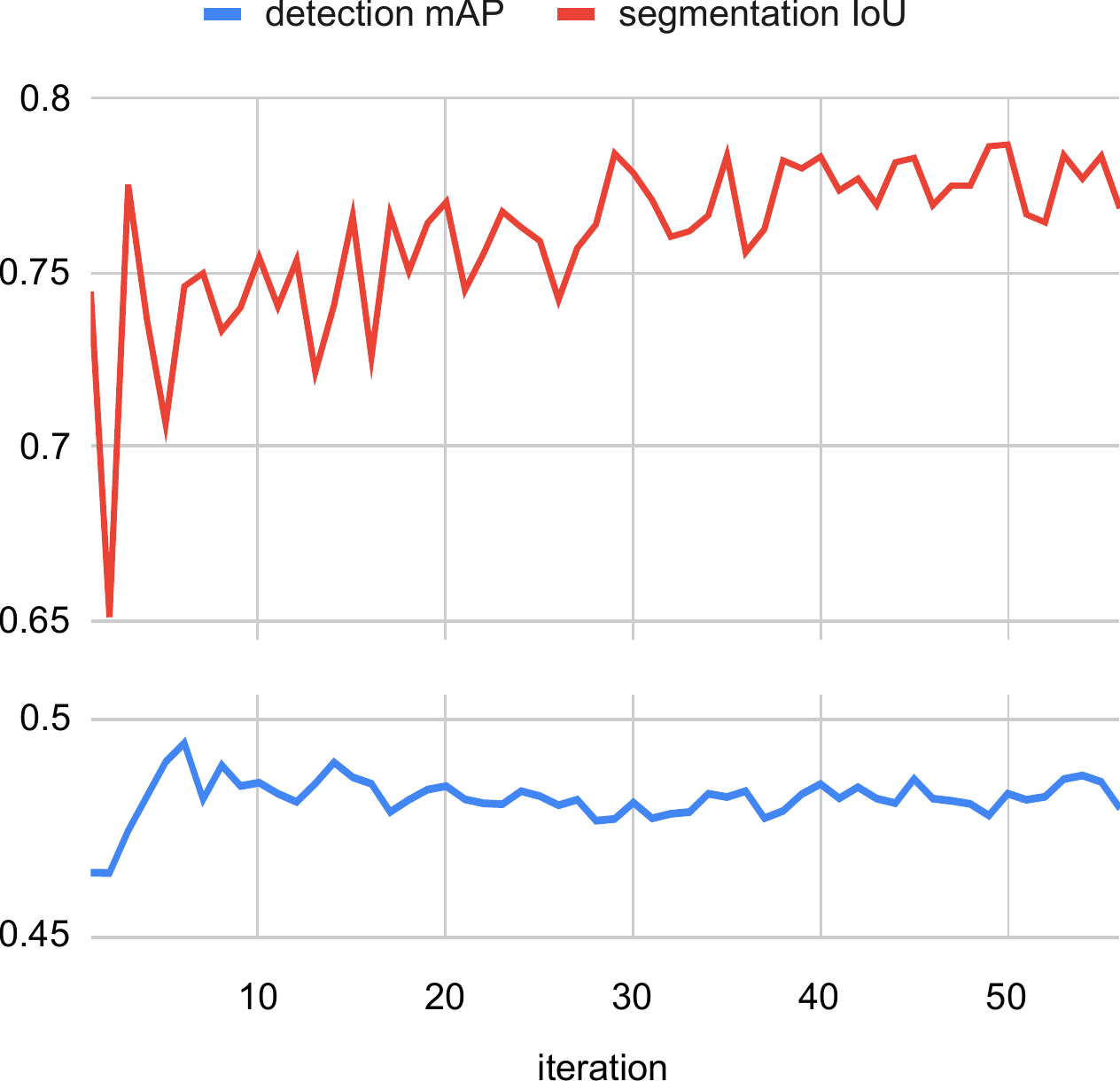}
\caption{Detection and segmentation performance over time.}
\label{fig:woodscape_metalearning_async_backprop_det_seg}
\end{subfigure}
\caption{Performance of meta-learning task weighting with asynchronous backpropagation method on WoodScape.}
\label{fig:woodscape_compromise_kpi}
\vspace{-20pt}
\end{figure}

{We accelerate optimization by adopting a relaxed version of network morphisms \cite{howard2019searching} that can be understood as a soft weight transfer that reuses the weights of the last best model as initialization for the offspring networks during the training. This enables to apply only a finetuning to the offspring networks and achieves a factor of four as speedup compared to from-scratch trainings.}

For each new configuration of hyper-parameters, we don't start from scratch, but instead train from the previously best model. In this way the number of epochs for each run can be effectively reduced (\eg, to 8 epochs for WoodScape dataset) by doing continuous finetuning while simultaneously tuning the hyperparameters.
% \ABc{I'm not sure what is actually happening from this last sentence. It would be helpful to clarify what is actually the dynamic weight transfer.}
One drawback of the meta-learning approach is increased computational cost as quite many partial trainings need to be performed to find the optimal solutions. This can lead to 4-6 times longer total runtimes compared to a single training. However, the ES optimization can well exploit multiple GPUs for speeding up. 
While training might be longer on particular known datasets, on the long run for new datasets for which typical training heuristics must be adapted and tested, meta-learning approaches clearly prove their effectiveness and utility.
All following experiments to new architectures and/or partially changed data can start from known parameter values to allow shorter optimization runtimes.

\subsection{Asynchronous backpropagation with task weighting using Evolutionary Meta-learning}\label{sec:async_backprop}

% In order to balance the convergence speed 
% \ABc{Maybe we can use learning speed, opimization speed/pace instead of convergence speed? I find that convergence speed is a bit vague} 
In order to balance the pace of optimization 
of the tasks, one method can be to control the backpropagation frequency of the tasks \cite{teichmann2018multinet}. In this way, a task that converges faster is updated less often than a task that takes more time to learn. 
An implementation trick is to set the task loss weight to 0.0 for the epochs for which we want to slow down training for the fast task. 
\begin{align}
&\loss_{total}^{(t)} = w_{seg}^{(t)} \loss_{seg}^{(t)} + w_{det}^{(t)} \loss_{det}^{(t)} \\
&w_{seg}^{(t)} = 1.0 \text{ } \forall t \\
&w_{det}^{(t)} = 1.0 \text{ if } t \text{ mod } \nu_{det} == 0 \text{ else } 0.0 
\end{align}
with $\nu_{det}$ the update frequency of the detection task.
This frequency is optimized by the meta-learning method described in the previous section using a numeric variable in the range of 1 to 10, followed by a rounding operation to an integer. As the segmentation takes longer to converge, $\nu_{seg}$ is set to 1.
Note that this scheduling can be coupled with data for which annotations for only one of the tasks are available, \eg, segmentation.
% \ABc{I will try adding a paragraph in the afternoon trying to conclude with the main idea of this method. Practically we are learning a task-learning schedule.}

\begin{figure}[tpb]
\centering
\begin{subfigure}[t]{0.5\textwidth}
\centering
\includegraphics[width=0.7\linewidth,keepaspectratio]{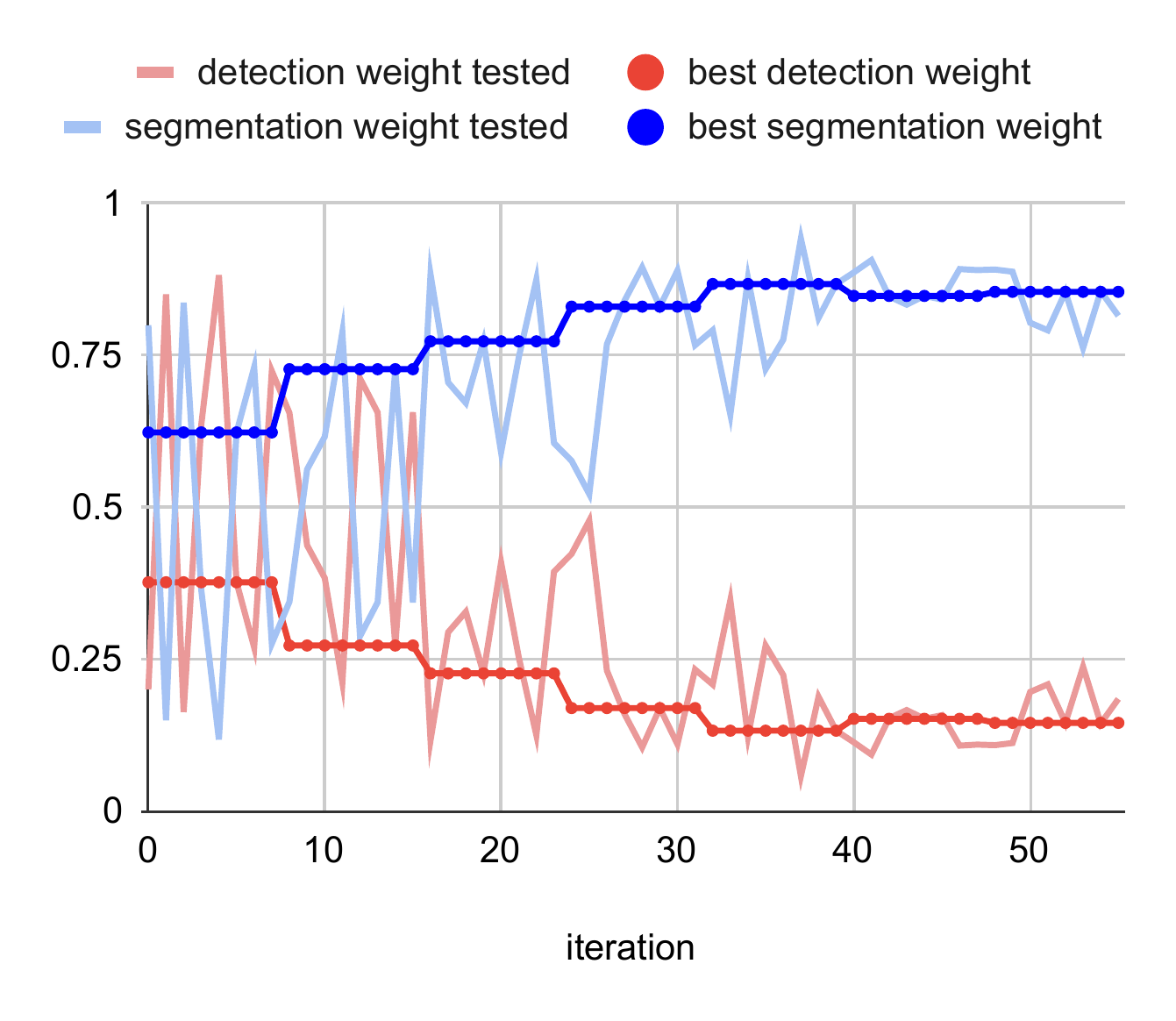}
\vspace{-0.2cm}
\caption{Tested task weights over time.}
\label{fig:woodscape_metalearning_async_backprop_det_loss_weight}
\end{subfigure}\hfill
\begin{subfigure}[t]{0.45\textwidth}
\centering
\includegraphics[width=0.7\linewidth,keepaspectratio]{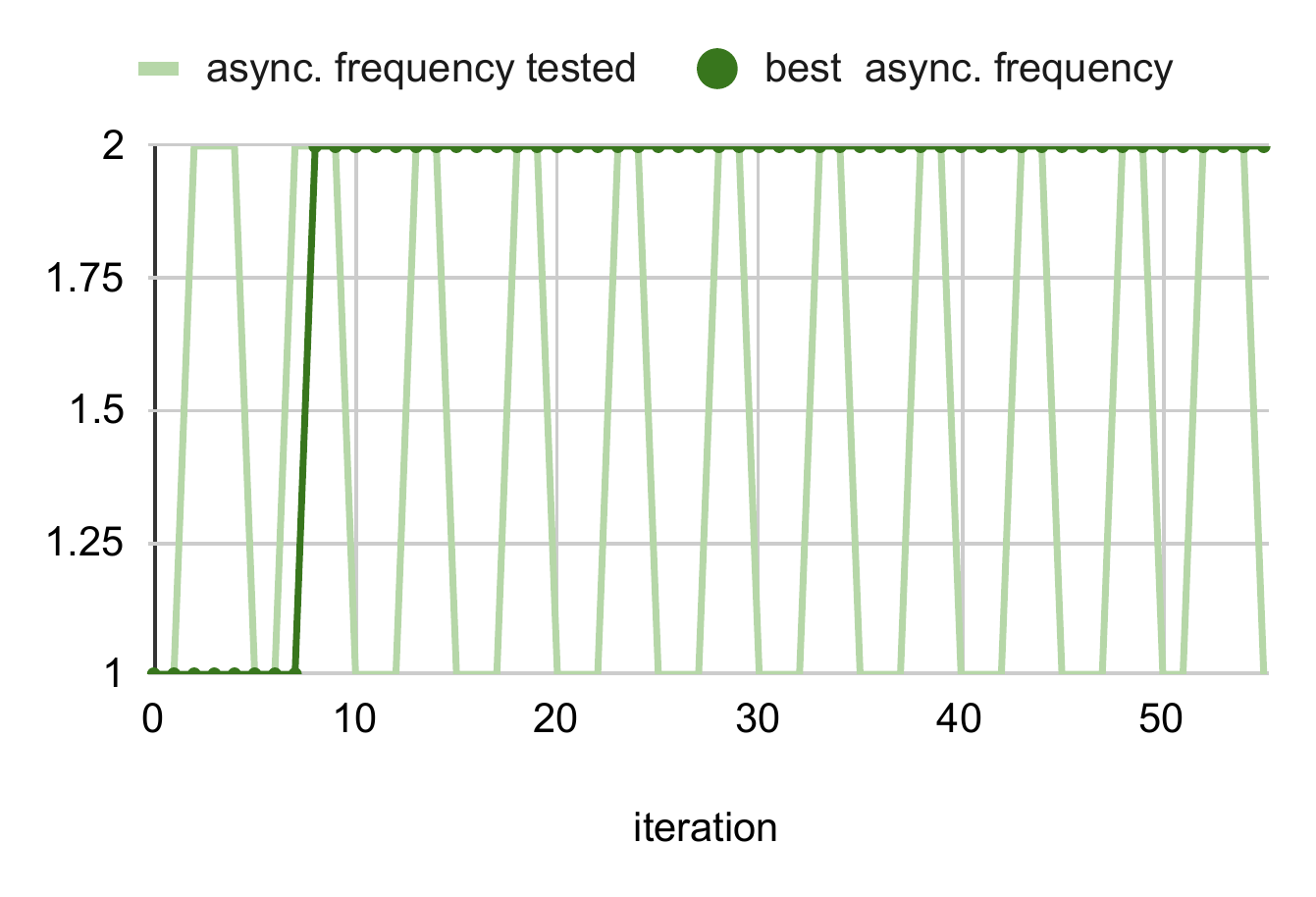}
\caption{Async. backpropagation parameter over time.}
\label{fig:woodscape_metalearning_async_backprop_async_freq}
\end{subfigure}
\caption{Task weights \& asynchronous frequency of detection task with Meta Asynchronous method on WoodScape dataset.}
\label{fig:woodscape_taskweights_asyncfreq}
\vspace{-20pt}
\end{figure}

\section{Results} \label{sec:results}

\begin{table*}[t]
\centering
%\vspace{-1pt}
\caption{Task weights and asynchronous backpropagation frequencies computed by several task-weighting methods.}
\vspace{-1pt}
\label{tab:task_weights}
\begin{adjustbox}{width=0.6\textwidth}
\begin{tabular}{llcccc}
\toprule
% \hline
&                                     & \begin{tabular}[c]{@{}l@{}}\textbf{No task} \\ \textbf{weighting}\end{tabular} & 

\begin{tabular}[c]{@{}l@{}}\textbf{Handcrafted} \\ \textbf{task weighting}\end{tabular} & 

\begin{tabular}[c]{@{}l@{}}\textbf{Meta-learning} \\ \textbf{task weighting}\end{tabular} & 

\begin{tabular}[c]{@{}c@{}}\textbf{Meta-learning task weighting} \\ \textbf{async backprop}\end{tabular} \\
% \hline
\midrule
\multirow{4}{*}{\textbf{KITTI}}      & $\bm{w_{seg}}$       & 1                          & 70                           & 0.8490                               & 0.9776                                                   \\
                                     & $\bm{w_{det}}$       & 1                          & 1                           & 0.1510                               & 0.0224                                                   \\
                                     & $\bm{\nu_{seg}}$ & -                          & -                                 & -                                    & 7                                                        \\
                                     & $\bm{\nu_{det}}$ & -                          & -                                 & -                                    & 1                                                        \\
% \hline
\midrule
\multirow{4}{*}{\textbf{Cityscapes}} & $\bm{w_{seg}}$       & 1                          & 40                           & 0.9478                               & 0.8692                                                   \\
                                     & $\bm{w_{det}}$       & 1                          & 1                           & 0.0522                               & 0.1308                                                   \\
                                     & $\bm{\nu_{seg}}$ & -                          & -                                 & -                                    & 1                                                        \\
                                     & $\bm{\nu_{det}}$ & -                          & -                                 & -                                    & 5                                                        \\
% \hline
\midrule
\multirow{4}{*}{\textbf{WoodScape}}  & $\bm{w_{seg}}$       & 1                          & 100                           & 0.9743                               & 0.8550                                                   \\
                                     & $\bm{w_{det}}$       & 1                          & 1                            & 0.0257                               & 0.1450                                                   \\
                                     & $\bm{\nu_{seg}}$ & -                          & -                                 & -                                    & 1                                                        \\
                                     & $\bm{\nu_{det}}$ & -                          & -                                 & -                                    & 2                                       \\
%  \hline
\bottomrule
\end{tabular}
\end{adjustbox}
\end{table*}

\begin{table*}[t]
%\vspace{-2pt}
\caption{Comparison of various task-weighting methods for  two-task network training.}
\vspace{-1pt}
\label{tab:accuracy}
\begin{adjustbox}{width=\textwidth}
\begin{tabular}{lllllllllll}
% \hline
\toprule
&                     & \begin{tabular}[c]{@{}l@{}}\textbf{No task} \\ \textbf{weighting}\end{tabular} & \begin{tabular}[c]{@{}l@{}}\textbf{Handcrafted} \\ \textbf{task weighting}\end{tabular} & \begin{tabular}[c]{@{}l@{}}\textbf{Dynamic task} \\ \textbf{loss scaling}\end{tabular} & \begin{tabular}[c]{@{}l@{}}\textbf{Uncertainty} \\ \textbf{weighting}\end{tabular} & \textbf{GradNorm} & \begin{tabular}[c]{@{}l@{}}\textbf{Geometric} \\ \textbf{loss}\end{tabular} & \begin{tabular}[c]{@{}l@{}}\textbf{Weight} \\ \textbf{learning}\end{tabular} & \begin{tabular}[c]{@{}l@{}}\textbf{Meta-learning} \\ \textbf{task weighting}\end{tabular} & \begin{tabular}[c]{@{}l@{}}\textbf{Meta-learning} \\ \textbf{task weighting} \\ \textbf{async backprop}\end{tabular} \\ 
% \hline
\midrule
\multirow{3}{*}{\textbf{KITTI}}      & \textbf{mAP}     & 0.6535                     & 0.6289                            & 0.1736                             & 0.6589                         & 0.6653            & 0.5677                  & 0.6727                   & 0.6974                               & \textbf{0.7260}                                          \\
                                     & \textbf{mIoU} & 0.8114                     & \textbf{0.8408}                   & 0.8079                             & 0.7974                         & 0.8080            & 0.8176                  & 0.8040                   & 0.8301                               & \textbf{0.8408}                                          \\
                                     & \textbf{G(mAP, mIoU)}            & 0.7282                     & 0.7272                            & 0.3745                             & 0.7248                         & 0.7332            & 0.6813                  & 0.7354                   & 0.7609                               & \textbf{0.7813}                                          \\
% \hline
\midrule
\multirow{3}{*}{\textbf{Cityscapes}} & \textbf{mAP}     & 0.2572                     & 0.2970                            & 0.2824                             & 0.2968                         & 0.2870            & 0.2900                  & 0.2972                   & 0.3091                               & \textbf{0.3177}                                          \\
                                     & \textbf{mIoU} & 0.6356                     & 0.5780                            & 0.5796                             & 0.5646                         & 0.5492            & \textbf{0.5819}         & 0.5573                   & 0.5812                               & 0.5815                                                   \\
                                     & \textbf{G(mAP,mIoU)}            & 0.4043                     & 0.4143                            & 0.4045                             & 0.4094                         & 0.3970            & 0.4108                  & 0.4070                   & 0.4239                               & \textbf{0.4298}                                             \\ 
% \hline
\midrule
\multirow{3}{*}{\textbf{WoodScape}}  & \textbf{mAP}     & 0.4643                     & 0.4438                            & 0.4557                             & 0.4525                         & 0.4511            & 0.4193                  & 0.4419                   & 0.4677                               & \textbf{0.4862}                                          \\
                                     & \textbf{mIoU} & 0.7180                     & 0.8107                            & 0.8118                             & 0.7806                         & 0.8155            & \textbf{0.8227}         & 0.8227                   & 0.8006                               & 0.7838                                                   \\
                                     & \textbf{G(mAP, mIoU)}            & 0.5774                     & 0.5998                            & 0.6082                             & 0.5943                         & 0.6065            & 0.5874                  & 0.6030                   & 0.6119                               & \textbf{0.6173}                                          \\ 
% \hline
\midrule
\textbf{}                     & \textbf{Average G(mAP, mIoU)}    & 0.5700                     & 0.5804                            & 0.4624                             & 0.5762                         & 0.5789            & 0.5598                  & 0.5818                   & 0.5989                               & \textbf{0.6095} \\                            
% \hline             
\bottomrule
\end{tabular}
\end{adjustbox}
\end{table*}

We conduct experiments on three automotive datasets. 
The proposed meta-learning method ($\mathcal{x}$\ref{sec:async_backprop}) outperforms the state of the art techniques \cite{Chen2018GradNormGN} and \cite{kendall2017multi} on all the three datasets  with a 3-4\% margin. 
% The method's only drawback is higher computational resources needed as multiple (shorter) trainings are performed. However, this can be justified with an increased performance and safety of the final ADAS application.
% \ABc{We can move this paragraph in the conclusions part of this section. I think it's better to first describe the experiments and protocol and then comment on results.}
We describe below the datasets we considered for this study, the evaluation protocol and metrics, and the results along some insights into the effect of our meta-learning method.

% \begin{figure}[t]
% \centering
% \includegraphics[width=\columnwidth]{fig/mtl_arch.pdf}
% \caption{Multi-task visual perception network architecture }
% \label{fig:multi-task}
% \end{figure}

\subsection{Datasets}

\textbf{KITTI}~\cite{Geiger2012CVPR} dataset for object detection consists of 7481 training images splitted into training and validation set. The dataset has bounding box annotations for cars, pedestrians and cyclists. For semantic segmentation task we have used \cite{krevso2016convolutional} that provided 445 images. Instead of 11 semantic classes we used only road, sidewalk and merged the other classes into void. %This not only helps to synchronize the classes with other two datasets but also 
This helps to simplify the analysis as semantic data is already highly imbalanced with much less data than for object detection. % and its important to balance the class distributions at overall pixel count. 

\begin{figure*}[htpb]
%\vspace{-2pt}
    \centering % <-- added
\begin{subfigure}{0.3\textwidth}
  \includegraphics[width=\linewidth]{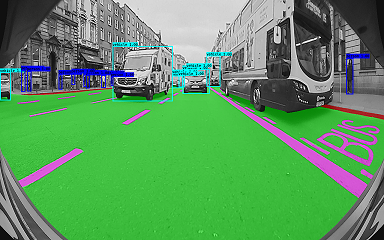}
  %\caption{image1}
  %\label{fig:1}
\end{subfigure}%\hfil % <-- added
\begin{subfigure}{0.3\textwidth}
  \includegraphics[width=\linewidth]{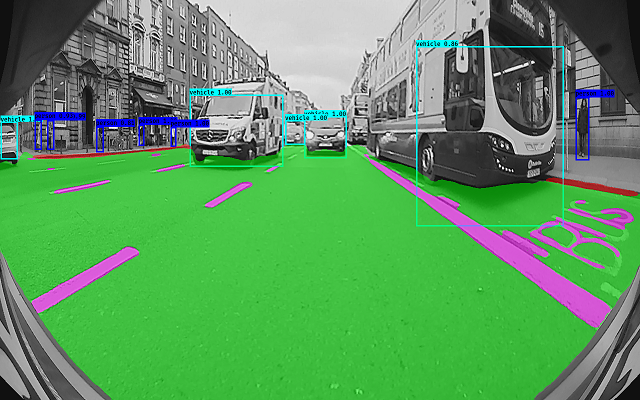}
  %\caption{image2}
  %\label{fig:2}
\end{subfigure}%\hfil % <-- added
\begin{subfigure}{0.3\textwidth}
  \includegraphics[width=\linewidth]{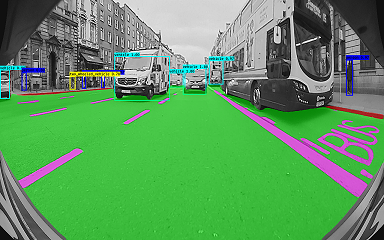}
  %\caption{image3}
  %\label{fig:3}
\end{subfigure}

% \medskip
% \begin{subfigure}{0.3\textwidth}
%   \includegraphics[width=\linewidth]{woodscape/gt/145814_RV_630.png}
%   %\caption{image4}
%   %\label{fig:4}
% \end{subfigure}\hfil % <-- added
% \begin{subfigure}{0.3\textwidth}
%   \includegraphics[width=\linewidth]{woodscape/prednoscale/145814_RV_630.png}
%   %\caption{image5}
%   %\label{fig:5}
% \end{subfigure}\hfil % <-- added
% \begin{subfigure}{0.3\textwidth}
%   \includegraphics[width=\linewidth]{woodscape/predasync/145814_RV_630.png}
%   %\caption{image6}
%   %\label{fig:6}
% \end{subfigure}

%\medskip
\begin{subfigure}{0.3\textwidth}
  \includegraphics[width=\linewidth]{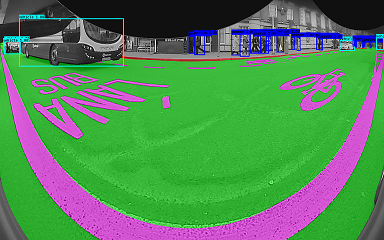}
  %\caption{image4}
  %\label{fig:4}
\end{subfigure}%\hfil % <-- added
\begin{subfigure}{0.3\textwidth}
  \includegraphics[width=\linewidth]{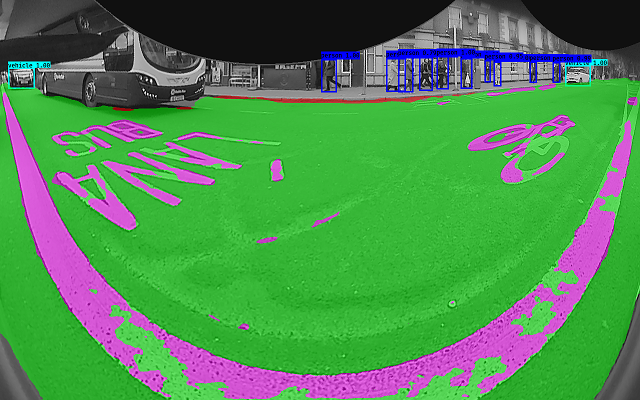}
  %\caption{image5}
  %\label{fig:5}
\end{subfigure}%\hfil % <-- added
\begin{subfigure}{0.3\textwidth}
  \includegraphics[width=\linewidth]{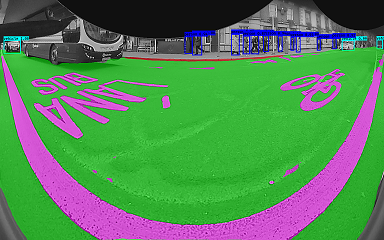}
  %\caption{image6}
  %\label{fig:6}
\end{subfigure}

% \medskip
% \begin{subfigure}{0.3\textwidth}
%   \includegraphics[width=\linewidth]{woodscape/gt/160910_MVL_450.png}
%   \caption{Groundtruth}
%   \label{fig:4}
% \end{subfigure}\hfil % <-- added
% \begin{subfigure}{0.3\textwidth}
%   \includegraphics[width=\linewidth]{woodscape/prednoscale/160910_MVL_450.png}
%   \caption{No task weighting}
%   \label{fig:5}
% \end{subfigure}\hfil % <-- added
% \begin{subfigure}{0.3\textwidth}
%   \includegraphics[width=\linewidth]{woodscape/predasync/160910_MVL_450.png}
%   \caption{Metalearning asynchronous backprop}
%   \label{fig:6}
% \end{subfigure}
% \caption{Results on Woodscape validation dataset.}
% \label{fig:woodscape_results}
% \end{figure*}

% \begin{figure*}[htb]
    %\centering % <-- added
% \begin{subfigure}{0.3\textwidth}
%   \includegraphics[width=\linewidth]{cityscapes/gt/image_00072.png}
%   %\caption{image4}
%   %\label{fig:4}
% \end{subfigure}\hfil % <-- added
% \begin{subfigure}{0.3\textwidth}
%   \includegraphics[width=\linewidth]{cityscapes/prednoscale/image_00072.png}
%   %\caption{image5}
%   %\label{fig:5}
% \end{subfigure}\hfil % <-- added
% \begin{subfigure}{0.3\textwidth}
%   \includegraphics[width=\linewidth]{latex/cityscapes/predasync/image_00072.png}
%   %\caption{image6}
%   %\label{fig:6}
% \end{subfigure}

\vspace{0.1cm}
\begin{subfigure}{0.3\textwidth}
  \includegraphics[width=\linewidth]{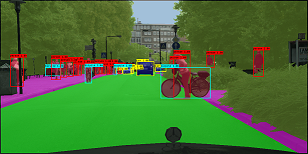}
  %\caption{image4}
  %\label{fig:4}
\end{subfigure}%\hfil % <-- added
\begin{subfigure}{0.3\textwidth}
  \includegraphics[width=\linewidth]{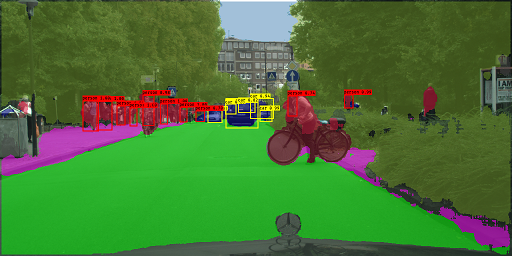}
  %\caption{image5}
  %\label{fig:5}
\end{subfigure}%\hfil % <-- added
\begin{subfigure}{0.3\textwidth}
  \includegraphics[width=\linewidth]{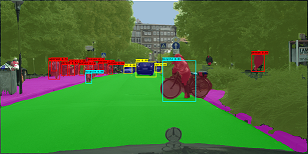}
  %\caption{image6}
  %\label{fig:6}
\end{subfigure}

%\medskip
\begin{subfigure}{0.3\textwidth}
  \includegraphics[width=\linewidth]{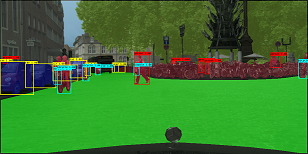}
  \caption{Groundtruth}
  \label{fig:4}
\end{subfigure}%\hfil % <-- added
\begin{subfigure}{0.3\textwidth}
  \includegraphics[width=\linewidth]{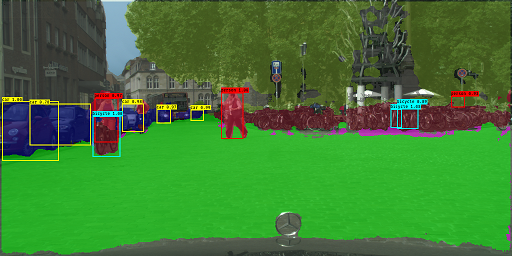}
  \caption{No task weighting}
  \label{fig:5}
\end{subfigure}%\hfil % <-- added
\begin{subfigure}{0.3\textwidth}
  \includegraphics[width=\linewidth]{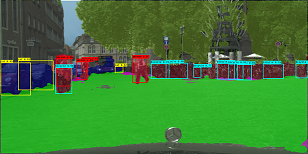}
  \caption{Metalearning asynchronous backprop}
  \label{fig:6}
\end{subfigure}
% \vspace{-7pt}
\caption{Quantitative results on WoodScape (top) and Cityscapes (bottom) validation dataset.}
\label{fig:cityscapes_results}
\end{figure*}

\textbf{Cityscapes} dataset~\cite{Cordts2016Cityscapes} consists of 5000 images with pixel level annotations. We extracted bounding boxes and semantic annotations from the provided polygon annotations. As the test data is not defined for bounding box regression, we have used at 60/20/20 split of the provided 5000 images for training, validation and testing.

\textbf{WoodScape}~\cite{yogamani2019woodscape} is an automotive fisheye dataset with annotations for multiple tasks like detection, segmentation and motion estimation. The dataset consists of 6K training, 2K validation and 2K test images. Instead of the 40 available semantic classes, we used only road, lanemarks, curb, person, two and four wheelers.  

%Similar to KITTI the proposed method has removed skewness towards segmentation performance.
%Similar to other datasets the existing task weighting methods favoured the segmentation task over the detection task.
{The proposed asynchronous meta-learning method will be particularly useful on unbalanced datasets like KITTI as it avoids overfitting of segmentation task on a small training set. Even for balanced datasets like Cityscapes and WoodScape, the method helps to regulate task convergence issues for detection task.}

% \ABc{We should add a few lines about the architectures.} 
\subsection{Implementation details}
% \subsubsection{Network architecture}
\textbf{Network architecture.} We have tested all the task weighting methods discussed in the previous section with a two-task network. We have designed a simple model which is suitable for low-power hardware. It consists of ResNet10 as a shared encoder, {a light version (10 layers) of residual networks with rapid convergence ~\cite{He2015}; YOLO style bounding box decoder ~\cite{redmon2016yolo9000} and FCN8 style semantic segmentation decoder ~\cite{long2015fully}. Our YOLO decoder composed of two convolutional layers is much simpler and faster to train than two-stage approach object detectors like Faster-RCNN.}
Figure~\ref{fig:multi-task} shows our network architecture. The Encoder head is pre-trained on ImageNet for all the experiments.

\textbf{Training settings.}
{
We used the loss from YOLO for object detection which is a combination of squared error losses and categorical cross-entropy loss for semantic segmentation.
For all the experiments, we train using ADAM optimizer with a learning rate of 0.0001 and we use a mini-batch size of 8. We train for 60 epochs on KITTI and Cityscapes and 50 epochs on WoodScape until convergence of the tasks.
Except for meta-learning experiments that take longer, we reduced the number of epochs to 30 on Cityscapes and 16 or 8 on WoodScape.
All the experiments run on a single GTX 1080Ti 11GB GPU except the meta-learning ones that exploit multiple GPUs. The training pipeline uses Tensorflow Keras framework.}

% \subsubsection{Meta-learning configuration}
\textbf{Meta-learning configuration.} {All the methods optimize two parameters simultaneously namely segmentation loss weight $w_{seg}$ and detection loss weight $w_{det}$. However, meta-learning is combined with asynchronous backpropagation and it optimizes two additional parameters namely asynchronous frequency for segmentation $\nu_{seg}$ and detection $\nu_{det}$.} %We optimized four parameters simultaneously, namely segmentation loss weight $w_{seg}$, detection loss weight $w_{det}$, asynchronous frequency for segmentation $\nu_{seg}$ and detection $\nu_{det}$. 
The variable ranges for the two task weights are $[0.1, 1000]$ for segmentation and $[0.1, 100]$ for detection as segmentation task usually profits from a higher weight due to longer convergence time. Table~\ref{tab:task_weights} shows the optimal values found out via optimization. The values represented are normalized between 0-1. The following optimization parameters for these experiments are determined empirically: size of initial population: 4, number of newly generated configuration: 4, number of parents per generated configuration: 2. 

\subsection{Evaluation metrics}
{All the experiments are evaluated using standard metrics on the validation set: \textbf{mAP} (mean Average Precision) ~\cite{DBLP:journals/corr/LinMBHPRDZ14} is used for object detection and \textbf{mIoU} (mean Intersection over Union) is used for semantic segmentation.
For object detection training and evaluation, small objects whose area is under 300 pixels are filtered as they are typically too far from the ego vehicle and hence unimportant.
Finally, we use geometric mean \textbf{G(mAP, mIoU)} (Table~\ref{tab:task_weights}) as the combined metric for the two tasks to enable comparison.}

\subsection{Results}
{We benchmarked the nine task weighting methods on the validation set of each dataset, results are detailed in Table~\ref{tab:accuracy}. 
The proposed meta-learning combined with asynchronous backpropagation method outperforms the others on the three datasets.
Table~\ref{tab:task_weights} shows the optimized task loss weights for meta-learning methods whose values are normalized ($\mathcal{x}$\ref{sec:metalearning} and $\mathcal{x}$\ref{sec:async_backprop}). We added the static weights computed by the handcraft task weighting method as a reference for comparison as it gives an order of magnitude of scale between segmentation and detection loss. Detection loss is x70 bigger than segmentation loss on KITTI, x40 on Cityscapes and x100 on WoodScape. All the methods in general try to weigh segmentation more than detection to compensate for this imbalance as seen in Table~\ref{tab:task_weights}.
Asynchronous version works better because it slowed down segmentation training on KITTI by a factor x7 to avoid overfitting on the small training set. 
Even for balanced datasets (same number of samples for both tasks) where segmentation does not show overfitting, detection usually converges faster than segmentation and might overfit. In this case, the asynchronous version slows down detection training by a factor x5 on Cityscapes and x2 on WoodScape as seen in Table~\ref{tab:task_weights}.}

\textbf{Insights into the meta-learning method}
In order to understand the optimization of the proposed meta-learning approach, some insights into the results on the WoodScape dataset are discussed in the following. Figure \ref{fig:woodscape_metalearning_async_backprop_main_metric} shows the target  metric over tested configurations for the optimization of task weights and the asynchronous backpropagation parameter. From initially low values, a slow but steady increase is observed. The best configuration is obtained after 44 iterations.

% \begin{figure}[htpb]
% \centering
% \includegraphics[width=0.45\textwidth]{fig/metalearn_mainmetric.pdf}
% \caption{Target metric over time for the Metalearning method on the Woodscape dataset with asynchronous backpropagation. The shown metric is the geometric mean of mIoU and mAP.}
% \label{fig:woodscape_metalearning_async_backprop_main_metric}
% \end{figure}
Figure \ref{fig:woodscape_metalearning_async_backprop_det_seg} shows the progression of the metrics of the two tasks during optimization. The segmentation performance is initially low and noisy and then steadily increases. The detection metric reaches it maximum early then degrades slightly to allow a compromise in favor of the segmentation towards the end of the optimization.
% \begin{figure}[htpb]
% \centering
% \includegraphics[width=0.4\textwidth]{fig/graphmetalerning_det_seg.pdf}
% \caption{Detection and segmentation performance over time for the Metalearning method on the Woodscape dataset with asynchronous backpropagation.}
% \label{fig:woodscape_metalearning_async_backprop_det_seg}
% \end{figure}
Figure \ref{fig:woodscape_metalearning_async_backprop_det_loss_weight} shows the progression of the task loss weights, and Figure \ref{fig:woodscape_metalearning_async_backprop_async_freq} the progress of the asynchronous backpropagation parameter over time during optimization. Figure \ref{fig:cityscapes_results} contains qualitative examples on WoodScape and Cityscapes validation dataset demonstrating improvements by the proposed method.

% \begin{figure}[htpb]
% \centering
% \includegraphics[width=0.42\textwidth]{fig/metalearn_weights2.pdf}
% \caption{Tested task weights over time for the Metalearning method on the Woodscape dataset with asynchronous backpropagation.}
% \label{fig:woodscape_metalearning_async_backprop_det_loss_weight}
% \end{figure}
% \begin{figure}[htpb]
% \centering
% \includegraphics[width=0.4\textwidth]{fig/async_freq_metalearn}
% \caption{Async. backpropagation parameter over time for the Metalearning method on the Woodscape dataset with asynchronous backpropagation.}
% \label{fig:woodscape_metalearning_async_backprop_async_freq}
% \end{figure}

%------------------------------------------------------------------------
\section{Conclusion} \label{sec:conc}
Multi-task learning provides promising performances in autonomous driving applications and is key in enabling efficient implementations at a system level. In this work, we take a closer look at this paradigm, which albeit popular has been rarely benchmarked across the same range of tasks and datasets. We thus evaluate nine different weighting strategies for finding the optimal method of training an efficient two-task model. We further propose two novel methods for learning the optimal weights during training: an adaptive one and one based on metalearning. Our proposed method outperforms state-of-the-art approaches by $3\%$ in compromise value.
In future work, we intend to extend our benchmarking to additional tasks, \emph{e.g.} on the wide range of tasks from the WoodScape dataset~\cite{yogamani2019woodscape}. 
% Multi-task learning can provide a promising performance in autonomous driving applications and is key in enabling an efficient implementation at a system level. In this paper, we evaluated nine different weighting strategies for finding the optimal method of training an efficient two task model.
% We proposed a novel method which beats the state of the art by 3\% in compromise value. In future work, we plan to evaluate the methods for more tasks in the WoodScape dataset.

%\newpage
{\small
\bibliographystyle{ieee}
\bibliography{egbib}
}

\end{document}